\documentclass{article}
\usepackage{amsmath,epsfig,multirow}
\usepackage[preprint]{spconfa4}
\usepackage{xcolor}
\usepackage{cite}

\usepackage{amssymb}
\usepackage{enumitem}

\usepackage{float}
\usepackage{xr}
\usepackage{makecell}
\usepackage{hyperref}
\usepackage{marvosym}

\let\OLDthebibliography\thebibliography
\renewcommand\thebibliography[1]{
  \OLDthebibliography{#1}
  \setlength{\parskip}{0pt}
  \setlength{\itemsep}{0pt plus 0.3ex}
}

\newcolumntype{P}[1]{>{\centering\arraybackslash}p{#1}}

\begin{document}\sloppy

% Example definitions.
% --------------------
\def\x{{\mathbf x}}
\def\L{{\cal L}}

% Title.
% ------
\title{Deep Geometry Post-Processing for Decompressed Point Clouds}
%
% Address.
% ---------------
\name{Xiaoqing Fan\textsuperscript{1}, Ge Li \Letter\textsuperscript{1}, Dingquan Li\textsuperscript{2}, Yurui Ren\textsuperscript{1}, Wei Gao\textsuperscript{1}, Thomas H. Li\textsuperscript{1,3}}
% \address{\textsuperscript{1}School of Electronic and Computer Engineering, Peking University Shenzhen Graduate School, China. \\
\address{\textsuperscript{1}Peking University Shenzhen Graduate School, China. \\
\textsuperscript{2}Peng Cheng Laboratory, Shenzhen, China.  \textsuperscript{3}AIIT, Peking University, Hangzhou, China.}

\maketitle

\begin{abstract}
Point cloud compression plays a crucial role in reducing the huge cost of data storage and transmission.
However, distortions can be introduced into the decompressed point clouds due to quantization.
In this paper, we propose a novel learning-based post-processing method to enhance the decompressed point clouds.
Specifically, a voxelized point cloud is first divided into small cubes.
Then, a 3D convolutional network is proposed to predict the occupancy probability for each location of a cube.
We leverage both local and global contexts by generating multi-scale probabilities.
These probabilities are progressively summed to predict the results in a coarse-to-fine manner.
Finally, we obtain the geometry-refined point clouds based on the predicted probabilities.
Different from previous methods, we deal with decompressed point clouds with huge variety of distortions using a single model.
Experimental results show that the proposed method can significantly improve the quality of the decompressed point clouds, achieving 9.30dB BDPSNR gain on three representative datasets on average.
\end{abstract}
%
% \footnote{
% Corresponding Author: Ge Li(geli@ece.pku.edu.cn) \\
% This study was supported by National Natural Science Foundation of China (No. 62172021), Shenzhen Fundamental Research Program(GXWD20201231165807007-20200806163656003).
% }

\section{Introduction}
\label{sec:intro}

\begin{figure}[t]
\begin{center}
\includegraphics[width=.9\linewidth]{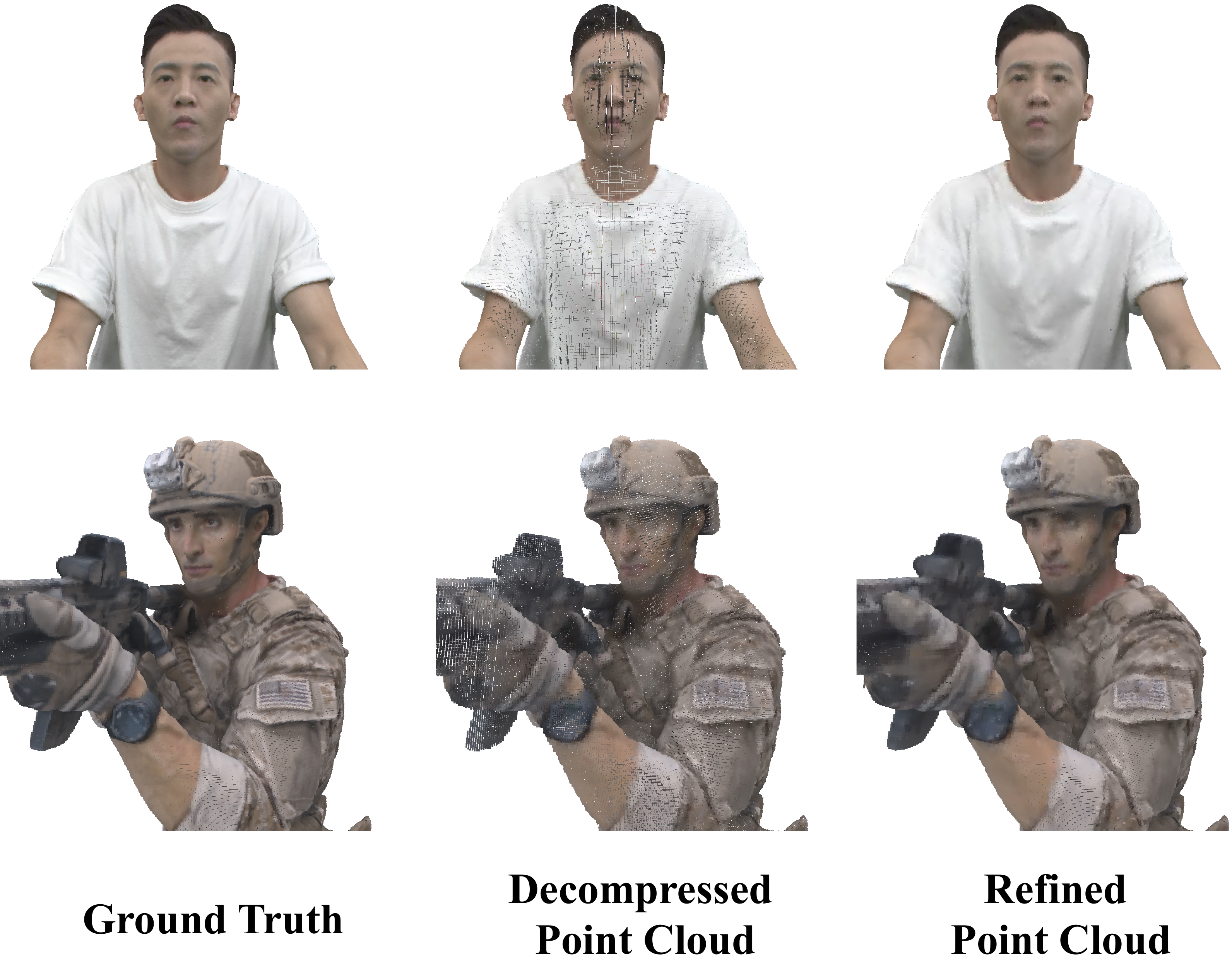}
\end{center}
   \caption{From left to right: the ground truth point clouds, the decompressed point clouds obtained by G-PCC~\cite{tmc13}, and the refined point clouds obtained by our model. Our model is able to significantly improve the geometry quality of the decompressed point clouds.}
 
\label{fig:sample}
\end{figure}

Point clouds are 3D point sets containing both geometry coordinates and associated attributes, which can describe scenes correctly and stereoscopically. 
Recently, point clouds have been increasingly applied to multimedia industries with immersion, interaction, and realism requirements. 
A point cloud can contain millions of points, which brings great difficulty for efficient storage and transmission.
Therefore, compression techniques are used to reduce the redundant information in point clouds.
However, most of the current compression schemes~\cite{schnabel2006octree,huang2008generic,kammerl2012real} apply quantization operation to reduce the point cloud size.  
During the quantization process, spatially adjacent points are merged into one point, which results in lower Level-of-Detail (LoD) decompressed point clouds.
As shown in Fig.~\ref{fig:sample}, the decompressed point clouds provide poorer visual quality compared with the original ones.

Post-processing techniques aim to refine the decompressed point clouds.
% an effective way to further refine the lower LoD decompressed point clouds.
Recently, some methods up-sample the decompressed point clouds to obtain higher LoD point clouds.
Borges~\textit{et al.}~\cite{borges2021fractional} super-resolve voxelized point clouds based on lookup tables.
This method effectively improves the quality of point clouds at different bit rates.
However, it requires constructing a lookup table for each point cloud, which reduces the efficiency of the method.
Akhtar~\textit{et al.}~\cite{akhtar2020point} apply a deep neural network to predict points lost during the quantization process. 
The upsampled results significantly improve the quality of input point clouds. 
However, this method sets a fixed up-sampling ratio for a model, which limits its application to tackle point clouds with arbitrary quantization steps. 

% Borges \textit{et al.}~\cite{borges2021fractional} present a method to super-resolve voxelized point clouds based on lookup tables, which can effectively improve the quality of point clouds at different bit rates.
% However, it needs constructing a lookup table for each point cloud.
% Akhtar \textit{et al.}~\cite{akhtar2020point} apply a sparse convolutional network to predict points lost during the quantization process. 
% The upsampled results significantly improve the quality of the input point clouds. 
% However, both these two approaches are short of the generalization capability.
% The former method needs constructing a lookup table for each point cloud, while the latter method requires training different models for different quantization steps, which greatly reduce their efficiency and limit their applications.

To deal with these limitations, we propose a learning-based post-processing model to refine decompressed point clouds.
We propose a 3D convolutional network to predict the distributions of original point clouds, which can handle decompressed point clouds with varying degrees of distortions.
% Since different distortions lead to different 
% 因为不同程度的失真会导致不同的LoD，因此我们使用三维卷机的方法来预测
Specifically, the decompressed point clouds are first voxelized and split into small cubes.
Then, the proposed convolutional network is used to refine the cubes by predicting the occupancy probability for each voxel.
We generate the final probabilities in a coarse-to-fine manner, where multi-scale probability cubes are sequentially generated and summed to predict the final probabilities.
% The final probabilities are computed by upsampling and summing the contributions of different scales.
Two types of determination strategies are used to screen out the refined point clouds from the predicted probabilities. 
% In addition to setting a fixed threshold, an adaptive threshold strategy based on the ground truth point numbers of cubes is also adopted for determination.
In addition to setting a fixed threshold, an adaptive threshold strategy is also adopted for determination based on the point numbers of the ground-truth cubes.
% based on the point numbers in the ground truth cubes
% Due to our unified predicting approach, once trained, our model is flexibly applied to refine point clouds at diverse bit rates.
% We show that our model can significantly improve the quality of the decomposed point clouds.
% Besides, we compare our proposed model with several state-of-the-art methods, and the experimental results of both R-D performance and BDPSNR demonstrate the superiority of our model.

Experimental results reveal that the proposed method significantly improves the quality of the decompressed point clouds, by gaining 9.30dB BDPSNR~\cite{bjontegaard2001calculation} using D1 (point-to-point) distance~\cite{mpeg2021ctc}. Comparison results show that our method achieves much higher quality point clouds than several state-of-the-art post-processing methods. 
% Besides, we have also reported that our method could provide much better visual quality.

In conclusion, our contributions are as follows: 
\vspace{-1.8mm}
 \begin{itemize}
  \setlength\itemsep{-0.2mm}
  \item We propose a deep neural network to enhance the quality of the decomposed point clouds. Our model is able to deal with point clouds with large varying degrees of distortions.
  \item We leverage both local and global contexts by generating probabilities in a coarse-to-fine manner. The ablation study proves the efficiency of this operation.
  % which is proven to be effective for the final accurate prediction results.
  \item Compared with several state-of-the-art methods, the proposed model demonstrates significant improvements on three benchmark datasets.
\end{itemize}

\section{Related works}

% In this section, we briefly summarize the traditional point cloud compression background, 
% progress on deep point cloud compression, and some recent post-processing methods.

\subsection{Point Cloud Compression}

Point Cloud Compression (PCC) aims to reduce the redundant information in point clouds while preserving the quality of the original point clouds.
Under the support of the international standard organization Moving Picture Experts Group (MPEG)~\cite{schwarz2018emerging}, two popular solutions are proposed to achieve point cloud compression: Video-based PCC (V-PCC) and Geometry-based PCC (G-PCC). 
The former leverages successful 2D video compression technologies to compress the projection of the point cloud, 
% which includes occupancy map, geometry and texture images.
the latter considers points in 3D space and compresses the geometry and attributes separately.
For geometry coding, G-PCC uses octree or trisoup scheme to represent point clouds and encodes the structures. However, many points are lost after the quantization operation, which results in low-quality decompressed point clouds.
% In terms of attribute coding, G-PCC uses one of the three transforming methods -- the Region Adaptive Hierarchical Transform (RAHT), the Predicting Transform, the Lifting Transform -- to obtain the transformation coefficients.
% Then, the transformation coefficients are quantized and arithmetically encoded.
% In this paper, we further improve the geometry of the decompressed point clouds using a deep neural network.
% Here we consider the situation of lossy compression, after an octree pruning, the decoded point clouds will have less points compared with the original inputs. 
% At the same time, the attribute ability will also be influenced.
% Both of these two popular architectures are devoted to removing data redundancy, saving transmission bandwidth and obtaining high-quality point cloud in the decoder.
% for dynamic frames and Geometry-based compression PCC(G-PCC) for static frames. V-PCC projects point cloud into six reference surfaces and compress these surfaces through mature HEVC video encoder(264??). Besides, MPEG also has proposed Geometry-based compression PCC(G-PCC) for static frame. For geometry compression, G-PCC uses octree or trisoup(?) to represent point cloud.(how about process afterwards?) For color compression, G-PCC has adopted RAHT and Predlift to remove attribute redundancy. Both of these two popular architectures are devoted to removing data redundancy, saving transmission bandwidth and obtaining high-quality point cloud.
% Recently, the rising of deep learning has caught people's attention in plenty of areas, 
% In addition to the 

Recently, the success of deep learning has promoted the development of learning-based PCC methods. 
Huang~\textit{et al.}~\cite{huang20193d} use multilayer perceptrons (MLP) to achieve end-to-end lossy compression firstly.
This method achieves good results on datasets with limited points, such as ModelNet~\cite{wu20153d} and ShapeNet~\cite{chang2015shapenet}.
% However, it cannot be applied to real-world point clouds.
% which could get great performance in datasets with little points, such as ModelNet~\cite{wu20153d} and ShapeNet~\cite{chang2015shapenet}. 
% Inspired by the 2D convolution in image compression work~\cite{balle2018variational}, 
Quach~\textit{et al.}~\cite{quach2020improved} and Wang~\textit{et al.}~\cite{wang2021lossy} adopt 3D convolutional networks on voxelized point cloud blocks. 
By splitting point clouds into small cubes, these methods are able to compress point clouds with millions of points, such as 8i Voxelized Full Bodies (8iVFB)~\cite{d20178i} and Microsoft Voxelized Upper Bodies (MVUB)~\cite{loop2016microsoft}. 
% Afterwards, many end-to-end compression methods~\cite{wang2021multiscale,wen2020lossy} have been proposed which aim to achieve better Rate-Distortion (RD) performance.

% Besides, some works try to deal with learning-based attribute compression~\cite{quach2020folding},~\cite{sheng2021deep}. 
% Due to the huge diversity of the colors between different point clouds, the performance of these algorithms still has room for improvement.

\subsection{Post-processing Methods}

The post-processing task aims to improve the quality of the decompressed point clouds.
Some methods based on V-PCC attempt to refine the occupancy map~\cite{jia2021convolutional} or the near and far depth maps~\cite{jia2021deep} to improve the quality of the decompressed point clouds. 
% However, these methods can only deal with the V-PCC decompressed point clouds.
Besides, some other methods have been proposed to alleviate the problem of missing points caused by quantization in G-PCC.
% , some suitable schemes have been proposed to up-sample the sparse decompressed point clouds.
Borges~\textit{et al.}~\cite{borges2021fractional} first build lookup tables from self-similarities and then apply the obtained lookup tables to refine voxelized point clouds.
% However, this method needs to construct a novel lookup table for each point cloud, which greatly reduces its efficiency.
Akhtar~\textit{et al.}~\cite{akhtar2020point} use sparse convolutions to predict the points which are lost during the quantization process. 
The up-sampling results can effectively improve the quality of the decompressed point clouds.
However, the former method needs to construct a lookup table for each point cloud, while the latter requires training different models for different quantization steps.
% However, this method requires training a new model for each quantization step.
In contrast, our method successfully trains a single model for point clouds compressed with different bit rates, which greatly enlarges the application scenarios.
% The upsampling results of these two methods can significantly improve the ability of the decoded point clouds. However, the former can not build a lookup table which is suitable for various point clouds, and the latter  Here we propose a deep post-processing model to improve the geometry and attribute ability of the decoded point clouds, which can deal with plenty of point clouds at various bit rates.

% To deal with every quantization step, this model needs to be retrained, which spares so much time and storage space. Inspired by the method of predicting the points' appearance probability~\cite{quach2019learning, wang2021lossy}, we propose a with a U-Net like network could deal with point cloud in any bit rates. Besides, we also consider the attribute improvement based on G-PCC, which is beneficial for the decoder visual quality. In conclusion, our contributions are as follows:
 % generate a higher Level-of-Detail (LoD) point cloud easily.

% this method can only deal with the situation that the quantization step is a round number, it can not be effective at any bit rates as a result. Based on that, we devised a learning-based network which can improve the low LoD freely, without considering the size of quantization step. Besides, to the best of our knowledge, this is the first work to improve the attribute quality. Our contributions are as follows:
% related work: 两章，一个点云压缩，一个后处理

\externaldocument{experiments}

\section{Approach}

\begin{figure*}[t]
\begin{center}
\includegraphics[width=1\linewidth]{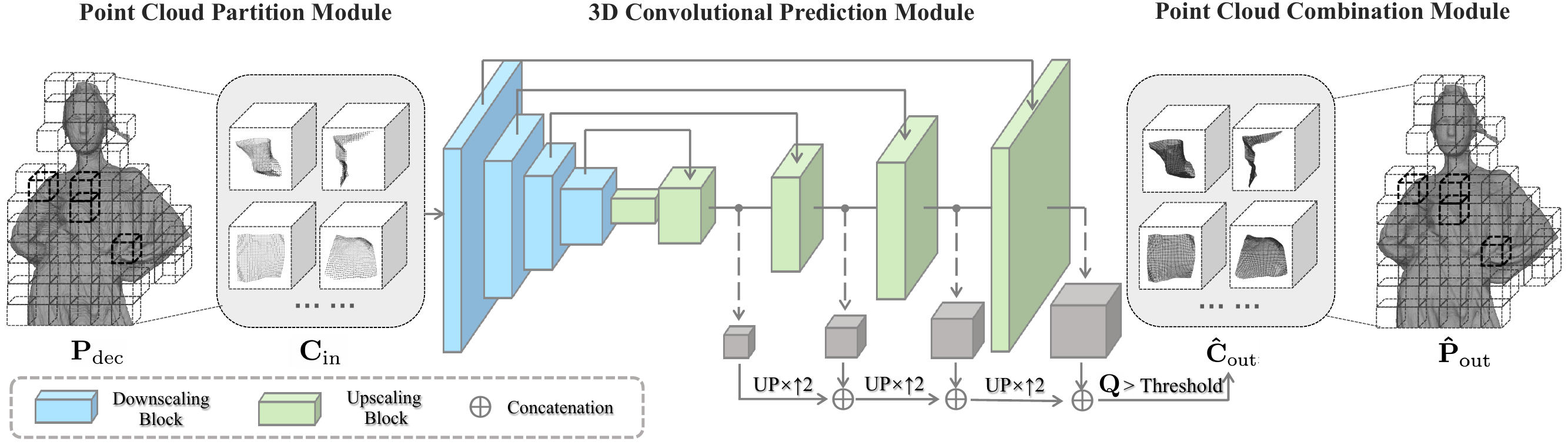}
\end{center}
   \caption{The architecture of the proposed model. We first split the point cloud into cubes. Then the neural network is used to predict the occupancy probability for each voxel. Finally, we obtain the refined point clouds based on the predicted probabilities.}
 
\label{fig:network}
\end{figure*}

In this section, we provide details of the proposed learning-based post-processing method. 
As shown in Fig.~\ref{fig:network}, 
the proposed model contains three modules: the Point Cloud Partition Module, the 3D Convolutional Prediction Module, and the Point Cloud Combination Module. 
The following content elaborates on the details.

% \subsection{Geometry Post-processing Network}

\subsection{Point Cloud Partition Module}

Point clouds collected from reality can have millions of points.
It is extremely difficult to refine all these points at one time. 
Therefore, partition methods are required to split point clouds for subsequent operations.
% 分块的操作是被需要的来预处理点云，以适合网络？处理？
In this paper, we follow the partition method described in~\cite{quach2020improved} and~\cite{wang2021lossy}. 
The decoded point clouds stored with disordered coordinates are first converted to a 3D volumetric representation $\mathbf{P}_{\mathrm{dec}}$ using \textit{voxelization}.
The voxelized point clouds are defined on regular 3D grids, whereas a voxel can be deemed as the 3D counterpart to the pixel in 2D.
Given that our model tackles the geometry information of the point clouds,
% network does not require the attribute information, 
we binarize $\mathbf{P}_{\mathrm{dec}}$ to obtain the geometry inputs, where occupied voxels are set to $1$, and the remaining voxels are set to $0$.

After obtaining $\mathbf{P}_{\mathrm{dec}}$, we divide it into small non-overlapped cubes $\mathbf{C}_{\mathrm{in}} \in \{0, 1 \}^{l \times w \times h}$.
Symbols $l$, $w$, and $h$ are the spatial sizes of cubes, respectively.
By converting the decoded point clouds into cubes, 3D geometry information of point clouds can be easily extracted by neural networks. 

\subsection{3D Convolutional Prediction Module}

% The 3D convolutional prediction module is responsible for generating the geometry-refined point clouds from the split cubes. 
% In the following, we first provide the details about the network architecture. 
% Then, we introduce the training function used to constrain our model. 
% Finally, we describe two determination strategies to screen out the final point clouds.

\vspace{0.1mm}
\noindent
\textbf{Multi-Scale Probability Prediction.}
After obtaining the split cubes $\mathbf{C}_{\mathrm{in}}$, our model refines these cubes by predicting the lost points caused by quantization and correcting the positions of existing points.
Point clouds compressed with different bit rates contain large varying degrees of distortion.
Therefore, to tackle point clouds with arbitrary distortion, the up-sampling ratio of the refinement network should not be explicitly limited.
To achieve this, we use a 3D convolutional neural network to generate the occupancy probability cubes $\mathbf{Q}\in \mathbb{R}^{l \times w \times h}$ from input cubes $\mathbf{C}_{\mathrm{in}}$.
By learning to predict the geometry distributions of the ground-truth cubes, our network can handle decompressed point clouds with varying degrees of distortion.
% can be easily used to model the distributions of point clouds compressed with arbitrary bit-rates.
% By mapping the inputs to the occupancy distributions of the ground-truth cubes, our model can tackle point clouds compressed with arbitrary bit-rates.
% By predicting the distributions of the ground-truth cubes, our model can tackle point clouds compressed with arbitrary bit-rates.

% 通过预测结果图像的分布，我们的模型可以轻易的适用于不同的码率输入

The architecture of the proposed network is shown in Fig.~\ref{fig:network}.
% We use a 3D convolutional network with a U-Net structure.
This network takes $\mathbf{C}_{\mathrm{in}}$ as inputs and generates the probability cubes $\mathbf{Q}$. Each value in $\mathbf{Q}$ indicates the occupancy probability of the corresponding voxel. The predicting process is described as follows:
\begin{eqnarray}
  \mathbf{Q} = \mathcal{G}(\mathbf{C}_{\mathrm{in}}),
\end{eqnarray}
where $\mathcal{G}$ denotes the 3D convolutional geometry network.
We design $\mathcal{G}$ using a U-Net structure.
The encoder consists of four downscaling blocks. 
Each block downsamples the inputs with a factor of $2$.
In the decoder, we use transposed convolutional layers with a stride of two to upscale the feature maps.
Skip connections are applied to concatenate the feature maps of the encoder and that of the decoder.
The concatenated feature maps are sent to the next upscaling block.
A sigmoid function is used as the activation function of the final layer, which enables obtaining values between $0$ and $1$.

We generate the final occupancy probability $\mathbf{Q}$ in a coarse-to-fine manner. 
Each decoder layer produces a probability cube with the corresponding scale. 
The obtained probability cube is then up-sampled using the nearest neighbor interpolation method.
We generate the final prediction by summing the contribution of each scale.
We show that the performance can be further improved by using such coarse-to-fine architecture in Sec.~\ref{sec:ablation}.

\vspace{0.1mm}
\noindent
\textbf{Loss Function.}
To train the model $\mathcal{G}$, we apply the same partition operation to the ground-truth point clouds and obtain binarized cubes $\mathbf{C}_{\mathrm{gt}}$. 
We calculate the cross-entropy loss $\mathcal{L}$ to minimize the difference between the refined occupancy probability $\mathbf{Q}$ and the ground-truth probability $\mathbf{C}_{\mathrm{gt}}$ as below:

\begin{eqnarray}
\mathcal{L} = -\frac{1}{N}\sum_{n=0}^{N}\big[c_n \text{log}(q_n)+(1-c_n)\text{log}(1-q_n)\big],
\end{eqnarray}
where $c_n$ and $q_n$ are the $n$-th voxel in $\mathbf{C}_{\mathrm{gt}}$ and $\mathbf{Q}$, respectively. Symbol $N = l \times w \times h$ denotes the number of voxels in a cube.

\vspace{0.1mm}
\noindent
\textbf{Determination Strategies.}
After obtaining $\mathbf{Q}$, the geometry-refined cubes $\mathbf{\hat{C}}_{\mathrm{out}}$ can be calculated using a suitable threshold $\sigma$. 
The choice of the threshold $\sigma$ is important since it directly determines the final prediction results. 
In this paper, we test two strategies including: 1) a fixed threshold $\sigma$ and 2) an adaptive $\sigma$ determined by the number of occupied voxels in $\mathbf{C}_{\mathrm{gt}}$. 
% A fixed threshold 
For the first strategy, 
we select the threshold with the highest performance on the training point clouds as our fixed threshold.
Here we use a fixed threshold $\sigma = 0.98$ for all experiments. 
Voxels with probabilities larger than $\sigma$ will be set as occupied.
Experiments show that by using the fixed threshold, our model is able to achieve state-of-the-art results.
For the second strategy, we use an adaptive $\sigma$, which selects top-k voxels to ensure the refined cubes $\mathbf{\hat{C}}_{\mathrm{out}}$ to have the same number of points as that of $\mathbf{C}_{\mathrm{gt}}$.
This setting requires additional bit streams to transmit the number of points in each ground-truth cube. 
The experiments in Sec.~\ref{sec:comparison} prove the transmission cost is negligible.
Meanwhile, this strategy can obtain additional RD performance gain compared with the fixed threshold.
It can still obtain even more RD performance gains with negligible increased bits.
% Experiments show that the negligible bandwidth consumptions always lead to additional performance gains. 
See Sec.~\ref{sec:comparison} for more details.

\subsection{Point Cloud Combination Module}

After obtaining $\mathbf{\hat{C}}_{\mathrm{out}}$, we combine these cubes to generate the geometry-refined point clouds $\mathbf{\hat{P}}_{\mathrm{out}}$. 
The combination operation first converts the relative coordinate $\mathbf{x}^{r}_{n}=(x^r_n, y^r_n, z^r_n)$ of each voxel in each cube to the absolute coordinate $\mathbf{x}_{n}=(x_n, y_n, z_n)$. Let us assume a cube with global index $\mathbf{i} = (i, j, k)$, the absolute coordinate is calculated as follows:
\begin{eqnarray}
  \mathbf{x}_{n} = \mathbf{i} \odot \mathbf{l} + \mathbf{x}^{r}_{n},
\end{eqnarray}
where $\odot$ is the element-wise multiplication. Symbol $\mathbf{l} = (l, w, h)$ is the spatial size of cubes. 
After obtaining the absolute coordinate $\mathbf{x}_{n}$, we copy the binary values in cubes $\mathbf{\hat{C}}_{\mathrm{out}}$ to the corresponding voxels in $\mathbf{\hat{P}}_{\mathrm{out}}$ to obtain the final refined point clouds.

\section{Experiments}

\subsection{Implementation Details}

\noindent
\textbf{Datasets.}
We use three datasets for training and evaluation.

\vspace{-2.5mm}
\begin{itemize}[leftmargin=*]
  \setlength\itemsep{-0.5mm}
\item \textit{8iVFB}~\cite{d20178i}. 
The 8i Voxelized Full Bodies is a dynamic point cloud dataset.
There are four sequences in the dataset. 
Each sequence contains around 300 frames recording the movements of a human subject.
The resolution is provided at 10-bit with a cube of $1024\times1024\times1024$ voxels.

\item \textit{MVUB}~\cite{loop2016microsoft}. 
The Microsoft Voxelized Upper Bodies is a dynamic voxelized point cloud dataset.
Five subjects are contained in this dataset.
The upper bodies of these subjects are captured. 
Each sequence contains around 250 frames.
% These point clouds 
% The surfaces of these point clouds are non-smooth and missing parts can be observed.
% \emph{i.e.} visible missing parts can be observed.
The resolutions are provided at 9-bit and 10-bit. We use 9-bit point clouds in the experiments.

\item \textit{ODHM}~\cite{yi2017owlii}. 
The Owlii Dynamic Human Mesh dataset contains four sequences. 
For each sequence, it contains around $400$ frames. 
The resolution is provided at 11-bit.

\end{itemize}

\vspace{-2.5mm}

\noindent
\textbf{Training Details.}
% The proposed model is trained in stages. 
% The geometry post-processing network $\mathcal{G}$ is first trained. 
% Then we train the attribute post-processing network $\mathcal{A}$ by using the geometry-refined point clouds obtained by $\mathcal{G}$.
% We use the longdress and loot sequences in the 8iVFB dataset for training.
The longdress and loot sequences in the 8iVFB dataset are used for training. 
We randomly select $60$ frames from the two sequences to construct the training set.
The latest version of MPEG-TMC13 (V14.0)~\cite{tmc13} is used to obtain the decompressed point clouds at different bit rates.
We train our model with decoded point clouds of multiple bit rates to improve its generalization.
The 3D block size $l \times w \times h$ is set to $64\times 64\times64$.
By default, we use a Adam optimizer with a learning rate $0.001$. The batch size is set to $64$.
% We set the fixed threshold as $\sigma=0.97$ for all experiment.

% implemented in PyTorch~\cite{paszke2019pytorch}. We train the model with NVIDIA Tesla P40 GPU using Adam optimizer for 40 epochs. By default, we use a learning rate of 1e-3 and a batch size of 16 for training. The 3D block size $l \times w \times h$ is set to 64×64×64 for both training and testing. The number of scalable layers is defined as 3 for geometry post-processing, 4 for color post-processing.
 %   the number of scalable layers, NL, was defined as 4. Stochastic gradient descent with the Adam algorithm [18] was used to train the model, using a learning rate of 10-4 and minibatches of eight 3D blocks at a time. 

\vspace{0.1mm}
\noindent
\textbf{Evaluation Details.}
We evaluate the performance of our proposed model following the MPEG common test conditions~\cite{mpeg2021ctc}. Except for the two training point cloud sequences, the other sequences of the above three datasets are used for testing.
The evaluation frame of each sequence is provided in Table~\ref{tab:geo}.
We use point-to-point error (D1)~\cite{mpeg2021ctc} to evaluate the geometry distortion.
% It calculates the distance between points in the reference point cloud and their nearest points in the testing point cloud.
It calculates the distance between points in a point cloud and the corresponding nearest points in another point cloud.
Both Peak Signal-to-Noise Ratio (PSNR) and Bjontegaard-Delta Peak Signal-to-Noise Rate (BDPSNR)~\cite{bjontegaard2001calculation} are used to report the D1 distortion values.
% The Peak Signal-to-Noise Ratio (PSNR) is used to report the D1 distortion values, which is defined as the peak signal over the symmetric distortion.
Higher scores indicate lower distortions.
% Besides, we employ Bjontegaard-delta peak signal-to-noise rate (BDPSNR)~\cite{bjontegaard2001calculation} to compare the quality gains at the equivalent bit rate.
% since our model introduces additional bit stream when using the adaptive threshold, we employ Bjontegaard-delta peak signal-to-noise rate (BDPSNR)~\cite{bjontegaard2001calculation} to compare the quality gains at the equivalent bit rate.

\begin{figure}[!t]
\centering
\includegraphics[width=1\linewidth]{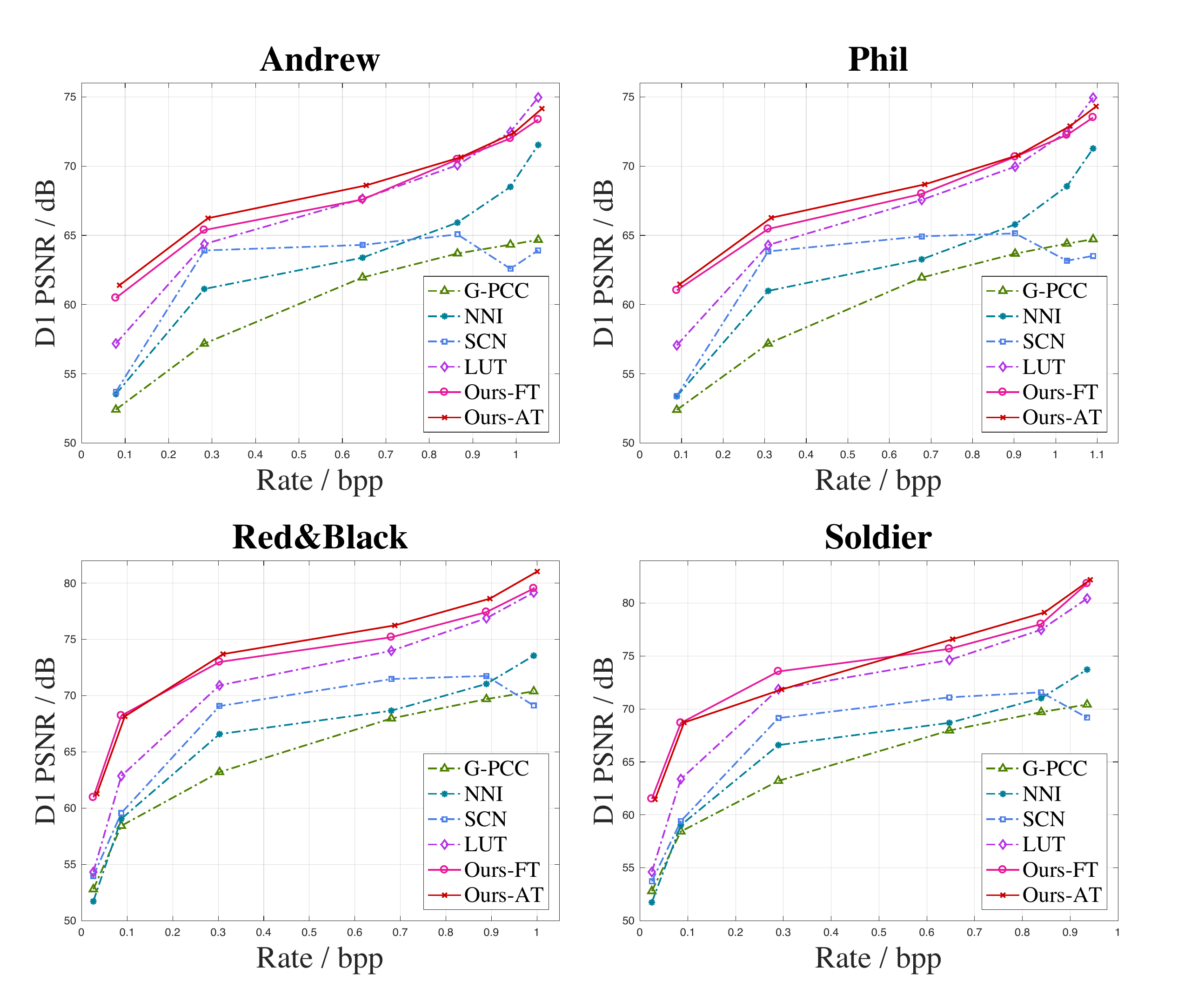}
\caption{Rate-distortion curves of different methods in several point clouds. For each curve, the horizontal axis represents the bit rate, while the vertical axis represents the PSNR score.}
\label{fig:RD_curve}
\end{figure} 

\begin{figure}[]
\begin{center}
\includegraphics[width=.95\linewidth]{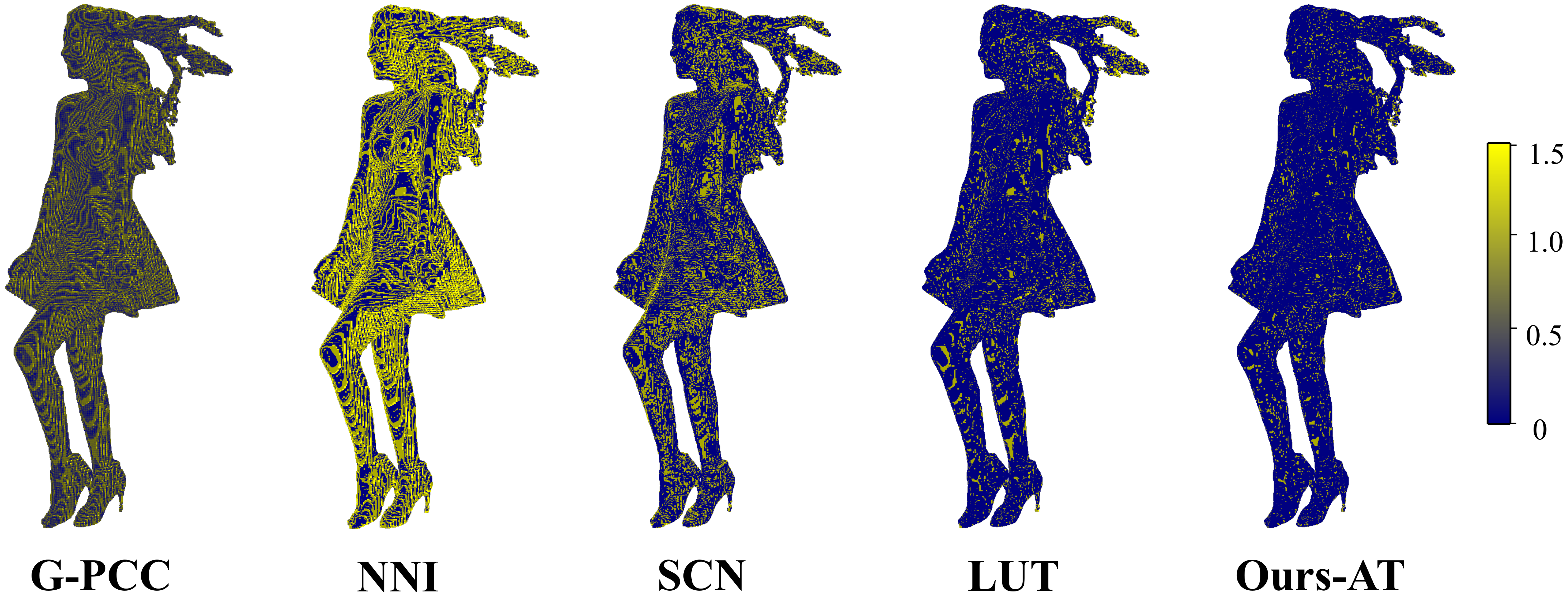}
\end{center}
   \caption{Visualization of point-to-point errors. The heat maps indicate the degrees of distortion. Blue represents the lowest distortion, while yellow represents the highest distortion.}
 
\label{fig:error_map}
\end{figure}

\begin{table*}[t]
% \begin{center}
\centering
\caption{Quantitative evaluation of different post-processing models. The BDPSNR gains against G-PCC (octree) are provided. We compare our model with several state-of-the-art methods including NNI, SCN~\cite{akhtar2020point}, and LUT~\cite{borges2021fractional}. Two determination strategies are evaluated: the fixed threshold $\sigma=0.98$ (Ours-FT) and the adaptive threshold (Ours-AT).} 
\label{tab:geo}
\vspace{2mm}
\resizebox{.95\textwidth}{!}{%
\begin{tabular}{p{1.2cm}P{2cm}P{1.2cm}P{1.2cm}P{1.2cm}P{1.2cm}P{1.2cm}P{1.2cm}P{1.3cm}P{1.3cm}} \Xhline{2\arrayrulewidth}
Dataset                & Point Cloud        & {\#}Frame                 & Depth     & Size        & NNI                  & SCN                & LUT    & Ours-FT         & Ours-AT  \\ \hline
\multirow{5}{*}{MVUB}  & andrew             & 0032                      & 9         & 281k        & 2.92                 & 4.18               & 6.47   & 7.64            & \textbf{8.47}             \\
                       & david              & 0032                      & 9         & 305k        & 2.77                 & 3.92               & 6.75   & \textbf{7.95}   & 7.19             \\
                       & phil               & 0032                      & 9         & 333k        & 2.79                 & 4.34               & 6.39   & 7.88            & \textbf{8.55}              \\
                       & ricardo            & 0032                      & 9         & 207k        & 2.93                 & 4.56               & 6.83   & 7.98            & \textbf{9.10}              \\
                       & sarah              & 0032                      & 9         & 301k        & 2.82                 & 4.10               & 6.65   & 7.84            & \textbf{8.68}    \\ \hline
\multirow{2}{*}{8iVFB} & redandblack        & 1550                      & 10        & 757k        & 1.27                 & 2.86               & 5.38   & 9.14            & \textbf{9.56}               \\
                       & soldier            & 0690                      & 10        & 1,089k      & 1.24                 & 2.69               & 5.99   & \textbf{9.64}   & 9.36                \\ \hline
\multirow{4}{*}{ODHM}  & basketball\_player & 0200                      & 11        & 2,925k      & 0.41                 & 2.38               & 5.66   & 10.41           & \textbf{10.85}               \\
                       & dancer             & 0001                      & 11        & 2,592k      & 0.59                 & 2.65               & 5.77   & 9.99            & \textbf{10.47}               \\
                       & exercise           & 0001                      & 11        & 2,391k      & 0.59                 & 2.76               & 5.65   & 10.01           & \textbf{10.47}               \\
                       & model              & 0001                      & 11        & 2,458k      & 0.77                 & 3.03               & 5.37   & 9.04            & \textbf{9.63}                \\ \hline
                       & \textbf{Average}   &                           &           & \multicolumn{1}{l}{} & 1.74        & 3.41               & 6.08   & 8.87            & \textbf{9.30}                \\ \Xhline{2\arrayrulewidth}
\end{tabular}
}
% \end{center}
\end{table*}

\subsection{Performance Comparison}
\label{sec:comparison}

In this subsection, we compare our model with several state-of-the-art methods: the Nearest-Neighbor Interpolation (NNI), Sparse Convolutional Network (SCN)~\cite{akhtar2020point}, and LookUp Table (LUT)~\cite{borges2021fractional}. 
NNI is a commonly used point cloud up-sampling method. 
It sets all children of each parent node as occupied.
Different from NNI, LUT up-samples point clouds with lookup tables calculated from self-similarities.
SCN is a learning-based method, which predicts the occupancy probabilities for the children of each voxel.
We retrain SCN using the same training datasets as ours for fair comparisons.
In SCN, we follow their original setting to train a new model for each bit rate. 
% On the contrary, our model only needs to be trained once for all bit rates.
Different from it, the advantage of ours is that we use a single model to deal with all bit rates.
% We use a fixed threshold $\sigma=0.98$ (Ours-FT) to determine the refined points.
\begin{table}
\centering
\caption{Quantitative evaluation of post-processing models. The BDPSNR gains against G-PCC (octree) are provided.} \label{tab:ablation}
\vspace{2mm}
\begin{tabular}{p{2cm}P{1cm}P{1cm}P{1cm}P{1cm}}  \Xhline{2\arrayrulewidth}
Dataset     & MVUB          & 8iVFB         & ODHM          & Average        \\ \hline
Baseline-FT & 6.51          & \textbf{9.45} & 9.42          & 8.10           \\
Ours-FT     & \textbf{7.86} & 9.39          & \textbf{9.86} & \textbf{8.87}  \\ \Xhline{2\arrayrulewidth}
\end{tabular}
\end{table}

\vspace{0.1mm}
\noindent
\textbf{Quantitative Comparison.} 
The quantitative comparison results are shown in Table~\ref{tab:geo}.
We provide the BDPSNR scores for each of post-processing models against G-PCC.
We also display the performance of our model with two determination strategies.
It can be seen that our method outperforms other methods at all test sets.
% We can achieve 9.30 BDPSNR gain on average. 
Although our model is trained on 10-bit point clouds, it still achieves good results on diverse datasets with different sizes.
The Rate-Distortion (R-D) curves are shown in Fig.~\ref{fig:RD_curve}.
These curves provide the performance of the models at different bit rates. 
It can be seen that our model can significantly improve the quality of the decompressed point clouds, especially at low bit rates.
Although LUT can obtain considerable performance gains at high bit rates, the performance decreases dramatically at low bit rates.

% The performance decreases sharply with the decrease of bit rate
% Compared with G-PCC, our model can significantly improve the performance of decoding point clouds at different bit rates. 
% Compared with SC method, our method performs better especially in the lower and higher bit rates, where our proposed method can achieve xx BDPSNR saving on average. 
% More results can be found in the supplementary material.

\vspace{0.1mm}
\noindent
\textbf{Qualitative Comparisons.} 
% We provide the generated results in Figure~\ref{fig:error_map}.
To intuitively show the distortions of the generated point clouds, we provide the error maps in Fig.~\ref{fig:error_map}. In these error maps, yellow and blue are used to indicate the highest and the lowest distortions, respectively.
It can be seen that our method can effectively reduce the errors between the decoded and the original point clouds.
% , which provides a better visual quality.

% 此外，通过error map，我们可以更加直观地看出我们的方法可以有效的减少解码点云与原始点云之间的误差，可以给人们带来更好的视觉效果。

% Table.~\ref{tab:geo} shows the average PSNR improvement cross different bit rates with the help of our method and Akahtar's method. Experiments show that our method can effectively improve the decoded point clouds in different data sets. We obtain an average gain of ??dB, which is superior to Akhtar's method. Besides, according to the error map Fig.~\ref{fig:??}, it is more intuitively observed that the refined point cloud of our method has less distortion. This also proves the superiority of our method.

% 表1是经过我们的方法和AA的方法对GPCC解码后的点云处理后，计算点云在六个码率点上提升的PSNR的平均值。相对于GPCC的方法，我们在所有数据集上平均有xxdB的提升。针对所有位宽和不同数据集条件下，我们的方法要远远好于AA的方法。根据图片xx，我们也可以更加直观地看出，我们的方法可以更加有效地提升点云质量，恢复的质量接近于原始点云。

% 从中，可以看出，我们的方法（fixed threshold）在三种数据集上都有着性能的提升。其中在11bit点云上的性能要相对较差，因为不同位宽点云之间的数据分布不同，仅仅使用固定阈值会对结果产生一定的影响，这也在后续章节中有所讨论。
% For color post-processing, to our knowledge, our work considers this task with deep learning for the first time. We compare our model with G-PCC (MPEG-TMC13 V14.0) here.

% 加数据集，从两个到MVUB到其他数据集（？
% 对于属性压缩，添加对比方法
% 计算减少的BDBR

\vspace{0.1mm}
\noindent
\textbf{Discussion on Determination Strategies.} 
% After obtaining the predicted occupancy probability, we screen out the final point clouds with threshold $\sigma$. 
We use two different determination strategies in this paper: the fixed threshold (Ours-FT) and the adaptive threshold (Ours-AT).
% The fixed threshold does not bring any bitstream overhead, while the adaptive threshold needs to transmit the number of points in each ground-truth cubes for selecting the top-k points in the final predictions.
The fixed threshold does not bring any bitstream overhead, while the adaptive one needs to transmit the number of points in each ground-truth cubes.
% The transmitted point numbers are compressed with a lossless codec: Lempel-Ziv-Markov chain-Algorithm (LZMA)~\cite{alakuijala2015comparison}.
We compressed the point numbers with a lossless codec: Lempel-Ziv-Markov chain-Algorithm (LZMA)~\cite{alakuijala2015comparison}.
According to the statistics, the adaptive strategy only needs to transmit a code stream less than 0.01bpp. 
Both RD curves in Fig.\ref{fig:RD_curve} and BDPSNR gains in Table~\ref{tab:geo} show that the adaptive strategy can always bring additional performance gain with negligible bits increase.

\begin{figure}[]
\begin{center}
\includegraphics[width=.9\linewidth]{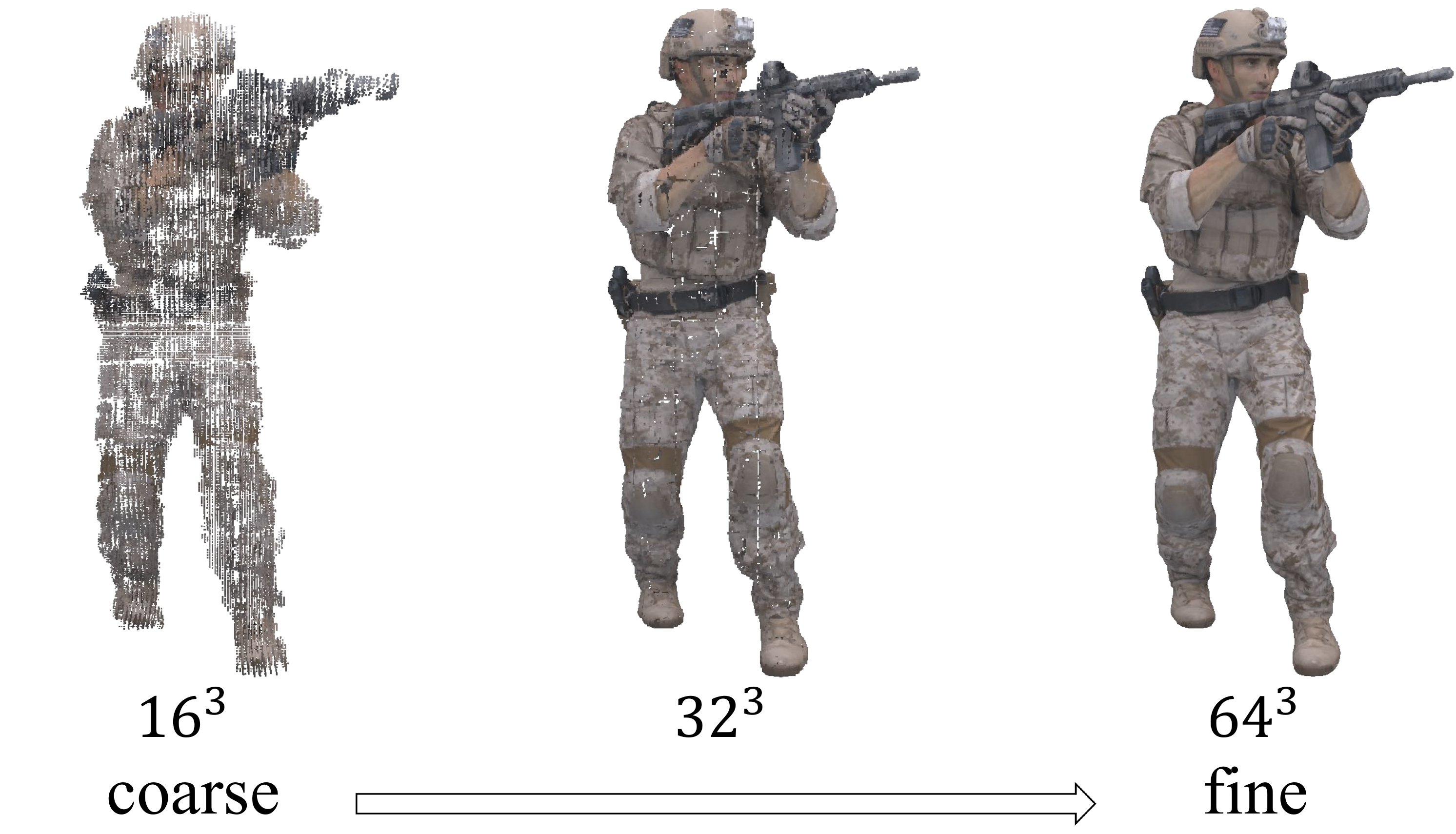}
\end{center}
   \caption{Visualization of the multi-scale probabilities predicted by our model. 
   Form left to right, we show the point clouds combined by cubes with resolutions $16^3$, $32^3$, and $64^3$.}
 
\label{fig:c2f}
\end{figure}  

\subsection{Ablation Study}
\label{sec:ablation}
We verify the effectiveness of the multi-scale probability prediction operation.
A baseline model is trained with a U-Net architecture by removing the coarse-to-fine structure from our model.
We train the baseline model using the same setting as ours.
The evaluation results are shown in Table~\ref{tab:ablation}.
It can be observed that our multi-scale probability prediction operation always brings additional performance gains.
We also show the intermediate results obtained from the multi-scale probability cubes in Fig.~\ref{fig:c2f}.
% It can be seen that the are generated in a coarse-to-fine process, which verify our 
A clear coarse-to-fine generation process can be observed, which verify our hypothesis.

% \subsection{Visualization}

\section{Conclusion}
We propose a deep post-processing network which can effectively enhance the quality of decompressed point clouds.
% A 3D convolutional network is proposed to predict the occupancy probabilities for each cube. 
% We first divide the complete voxelized point cloud into cubes of the same size. 
A 3D convolutional network is applied for predicting the occupancy probabilities of each position, in order to recover lost points caused by quantization. 
% A coarse-to-fine architecture is 
Moreover, we use a coarse-to-fine architecture to progressively generate the final predictions.
Different from previous methods, our model is successfully trained once to deal with point clouds with large varying degrees of distortion.
% Our model is able to deal with point clouds at diverse bit rates.
% Instead of training different models for each quantization step, our model is trained once to deal with decompressed point clouds at diverse bit rates.
Experimental results show that our model can generate geometry-refined point clouds, which significantly improves the visual quality.
Code is available at \href{https://github.com/fxqzb/Deep-Geometry-Post-Processing}{https://github.com/fxqzb/Deep-Geometry-Post-Processing}.

% \vspace{0.1mm}
\noindent
\textbf{Acknowledgment.}
This work was supported in part by National Natural Science Foundation of China (No. 62172021),
and in part by Shenzhen Fundamental Research Program (GXWD20201231165807007-20200806163656003).

\bibliographystyle{IEEEbib}
\bibliography{icme2022template}

\end{document}